\documentclass{article}
\usepackage{hyperref}
\usepackage{microtype}
\usepackage{graphicx}
\usepackage{subfigure}
\usepackage{float}
\usepackage{booktabs} 
\usepackage{amssymb}
\usepackage{amsfonts}
\usepackage{wrapfig}
\usepackage{adjustbox} 
\usepackage{bm}
\usepackage[table,xcdraw]{xcolor}  

\usepackage{hyperref}

\newcommand{\eg}{\textit{e.g.}}

\usepackage[accepted]{icml2025}

\usepackage{amsmath}
\usepackage{svg}
\usepackage{amssymb}
\usepackage{mathtools}
\usepackage{amsthm}
\usepackage{graphicx}
\usepackage{booktabs}
\usepackage{amssymb} 
\usepackage{pifont}  
\usepackage{array}   
\usepackage{enumitem}

\newcommand{\xmark}{\ding{55}} 

\usepackage[capitalize,noabbrev]{cleveref}

\theoremstyle{plain}

\theoremstyle{definition}

\theoremstyle{remark}

\usepackage[textsize=tiny]{todonotes}

\icmltitlerunning{Efficient Large Language Models Evaluation via Collaborative Filtering}

\definecolor{mygreen}{rgb}{0.1, 0.8, 0.1}  
\definecolor{myred}{rgb}{0.8, 0.1, 0.1}    

\begin{document}

\twocolumn[
\icmltitle{Efficient Evaluation of Large Language Models via Collaborative Filtering}



\icmlsetsymbol{equal}{*}
\begin{icmlauthorlist}
\icmlauthor{Xu-Xiang Zhong}{equal,nju,lab}
\icmlauthor{Chao Yi}{equal,nju,lab}
\icmlauthor{Han-Jia Ye}{nju,lab}
\end{icmlauthorlist}
\icmlaffiliation{nju}{School of Artificial Intelligence, Nanjing University}
\icmlaffiliation{lab}{National Key Laboratory for Novel Software Technology, Nanjing University}
\icmlcorrespondingauthor{Han-Jia Ye}{yehj@lamda.nju.edu.cn}
\icmlkeywords{Machine Learning, ICML}

\vskip 0.3in
]

\printAffiliationsAndNotice{\icmlEqualContribution}

\begin{abstract}
With the development of Large Language Models~(LLMs), numerous benchmarks have been proposed to measure and compare the capabilities of different LLMs.
However, evaluating LLMs is costly due to the large number of test instances and their slow inference speed.
In this paper, we aim to explore how to efficiently estimate a model's real performance on a given benchmark based on its evaluation results on a small number of instances sampled from the benchmark.
Inspired by Collaborative Filtering~(CF) in Recommendation Systems~(RS), we treat LLMs as users and test instances as items and propose a two-stage method.
In the first stage, we treat instance selection as recommending products to users to choose instances that can easily distinguish model performance.
In the second stage, we see performance prediction as rating prediction problem in RS to predict the target LLM's behavior on unselected instances.
Experiments on multiple LLMs and datasets imply that our method can accurately estimate the target model's performance while largely reducing its inference overhead.
\end{abstract}

\section{Introduction}\label{section:1}

Large Language Models~(LLMs) have garnered widespread attention, with numerous LLMs~\citep{qwen,llama,llama-2,chatglm} being released and rapidly iterated. 
Due to their powerful general capabilities, these LLMs are expected to perform a diverse and broad range of tasks~\citep{medicine, math, Dail-sql, ToolLLM}. 
To fairly assess and compare different LLMs' capabilities, many benchmarks for evaluating LLMs have emerged and developed continuously~\citep{agentbench,MMLU,CMMLU,agieval,HELM}.
To comprehensively and accurately assess the diverse capabilities of LLMs, LLM benchmarks typically contain a variety of tasks~(or scenarios), with each task corresponding to a sub-dataset. 
Furthermore, even within a specific task, such as code generation~\cite{code_x_glue}, there may be multiple programming languages, and each language has its own sub-dataset. 
So, a well-designed benchmark often contains many test instances.

\begin{figure}[t]
\centering  
\subfigure{
\includegraphics[width=\linewidth]{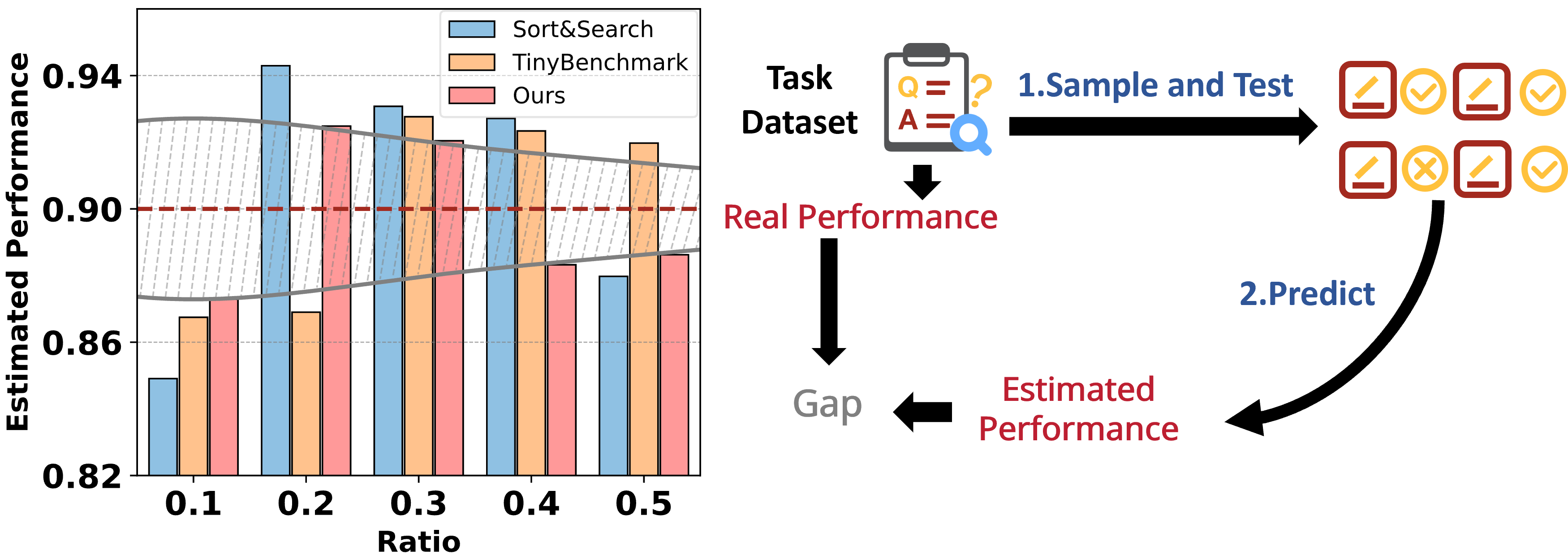}}
\vspace{-0.4cm}
\caption{\textbf{Comparison between Methods and Problem Setting.} On the left, the red dashed line represents the real performance of a new model, and the gray area indicates the gap between the estimated performance of our method and the real performance, which is smaller. On the right is the problem setup, where the goal is to extract a subset from each task, use the new model's evaluation on it to predict performance on each task and minimize the gap between estimated and real performance.}
\label{fig:1}
\vspace{-0.4cm}
\end{figure}

Given the large number of test instances and the relatively slow inference speed of LLMs, it is impractical to let LLMs infer on all instances to obtain their performance. 
For example, when selecting the best model for a target task from an LLM Zoo, evaluating all models on a benchmark would result in substantial time and computational overhead. 
Furthermore, inference with certain closed-source large models incurs fees, making it costly to run these models on a large volume of test samples. 
In summary, performing inference on all instances of the target benchmark for each LLM in a large model zoo is not viable in resource-constrained scenarios. 
We hope to accurately evaluate the capabilities of an LLM in one specific task at a low cost.

Some previous researches~\cite{lifelong1,lifelong_benchmark,tiny_benchmark} have proposed their solutions to this problem.
They only need to conduct inference on a small number of test instances to estimate the target models' performance. 
However, most of the previous researches focus only on the overall performance and ranking of LLMs across the entire benchmark, without considering the task-specific performance and ranking of LLMs on individual tasks within the benchmark.
Considering only overall performance cannot comprehensively and fairly evaluate the capabilities of large language models (LLMs), as different LLMs with similar overall performance and rankings may exhibit significant performance differences between tasks.
And in practical applications, we often focus more on an LLM's task-level capabilities rather than overall capabilities. 
Therefore, a low-cost method is urgently needed to evaluate the capabilities of large models in various tasks.

To meet the above need, we propose a two-stage method inspired by Collaborative Filtering~(CF) in recommendation systems~\cite{CB,CB2,survey_collaborative_fitering}. 
We apply collaborative filtering in both stages, treating LLMs as users and instances as items. 
Processes in two stages focus on determining the value of items for a new user. 
They utilize the interaction history of other users with items and the new user's interaction history with some items.
And \autoref{fig:1} is a conceptual diagram that integrates different tasks. We can observe that our method is closer to real performance compared to other methods.

In the first stage, we treat instance selection as recommending products to users. 
And we use CF to select a given number of instances, such as 10\% for each scenario according to the importance score to get the real evaluation results of the target model.
In the second stage, we view performance prediction as rating prediction problem in RS to predict the target LLM’s behavior on unselected instances.
Specifically, we predict performance using CF, based on the results of similar LLMs on remaing 90\% unselected instances and the results of the target LLM on selected instances and synthesized instances by optimal transport~\cite{ot}.
Our contributions are as follows:\vspace{-0.42cm}
\begin{itemize}[leftmargin=10pt]
\item We propose an efficient evaluation method based on the idea of collaborative filtering, which can efficiently give the performance of LLMs on different tasks.\vspace{-0.25cm}
\item We analyze the similarities and differences between efficient evaluation methods and Recommendation Systems~(RS), which inspire us to apply the methods of RS to address the efficient evaluating problem.\vspace{-0.25cm}
\item Experiments on benchmarks composed of various LLMs and tasks demonstrate the effectiveness of our method.
\end{itemize}

\section{Related Work}
\textbf{Efficient Benchmarking of PFMs.} With the development of Pre-trained Foundation Models (PFMs), multiple benchmarks are introduced to quantify PFM's abilities and compare different PFMs. 
However, the continuous growth in the size of models and datasets has increased evaluation costs. 
Some researchers focus on designing efficient benchmarks to accommodate the costs. 
\citet{efficient_benchmark} find that while diversity across datasets is crucial for evaluation in HELM~\citep{HELM}, the quantity of examples presently utilized is unnecessarily large. They also design a coarse-to-fine tournament algorithm to get LLM's ranking.
\citet{anchor_point} suggest grouping evaluation samples according to LLM's confidence in the correct class to accelerate evaluation processes for classification tasks.
TinyBenchmarks~\cite{tiny_benchmark} leverages models of educational assessments from psychometrics to accurately assess the capabilities of LLMs with a fraction of the test instances in standard benchmark datasets.
Lifelong Benchmarks~\citep{lifelong_benchmark} focuses on vision models. It proposes a method called Sort\&Search and this method leverages previous model predictions to get each test instance's difficulty and then ranks and selectively evaluates test instances.
In this paper, we design a new efficient benchmark method for LLMs based on the idea of collaborative filtering, which has superior performance.

\textbf{Data Selection for LLM.}
Some previous work has attempted to select training data for LLM during the training phase to reduce the impact of low-quality training instances on model performance and improve training speed and efficiency.
\citet{shapley_value} propose TS-DSHAPLEY to utilize Shapley Values to filter out harmful training data, thereby improving the performance after model fine-tuning.
\citet{DSIR} design Data Selection with Importance Resampling~(DSIR) to select a tailored subset from the pretraining dataset for a specific target instance distribution, aiming to maximize the performance of the pre-trained model while adhering to a fixed compute budget. DSIR estimates importance weights in a reduced feature space for tractability and selects data with importance resampling according to these weights.
\citet{DoReMi} leverage distributionally robust optimization~(DRO) to tune the domain weights without knowledge of downstream tasks. These domain weights decide the mixture proportions of pretraining data domains.
In this work, we primarily focus on instance selection during the testing phase of large language models.

\begin{figure*}[t]
    \centering  
    \includegraphics[width=\linewidth]{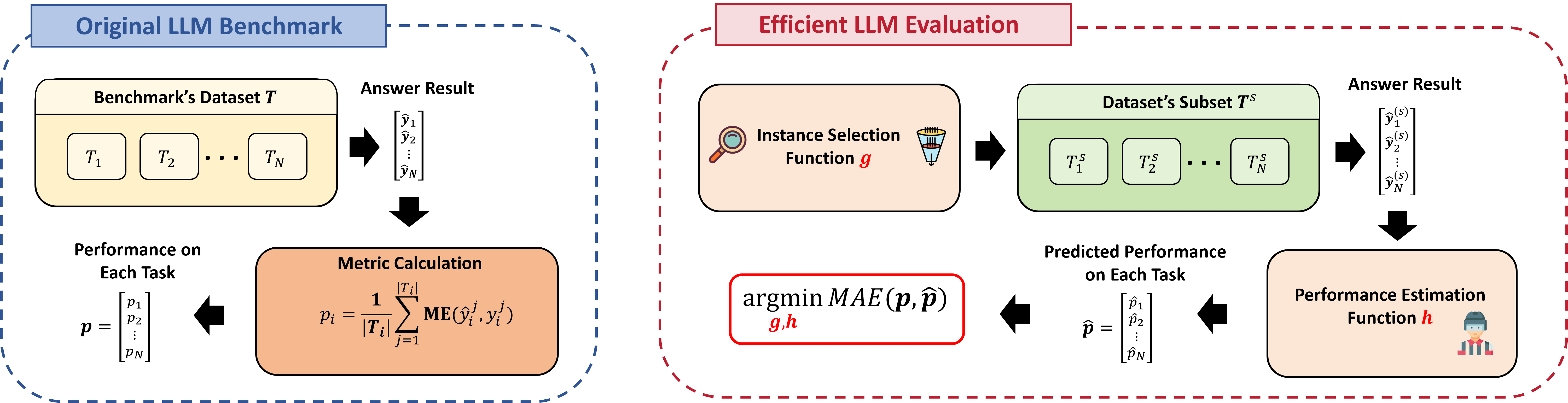} 
    \vspace{-0.4cm}
    \caption{\textbf{The Paradigms of Original and Efficient LLM Benchmark}. The left part illustrates the evaluation process of the Original LLM Benchmark. The right part shows the process of an efficient evaluating method, which consists of two main components: the Instance Selection Function $g$ and the Performance Estimation Function $h$. The goal of efficient evaluation methods is to design effective $g$ and $h$ to minimize the difference between real performance $\bm{p}$ and predicted performance $\hat{\bm{p}}$.}
    \label{fig:2}
    \vspace{-0.4cm}
\end{figure*}

\section{Preliminary}    
Here we will introduce the process of original LLM evaluation and efficient evaluation methods, as shown in \autoref{fig:2}. 
We will also present the evaluation metrics for comparing these methods and insights from a simple baseline.

\subsection{Evaluation for LLMs}\label{section:original_evaluation}
Assume there is an LLM Benchmark, whose dataset is $T=\{T_1,\cdots, T_N\}$ consisting of $N$ different tasks' datasets.
Here $T_i = \{(x_i^j, y_i^j)\}_{j=1}^{|T_i|}$ refers to the $i$-th task's dataset. 
$x_i^j$ and $y_i^j$ represent the instance text and label text of the $j$-th instance in $T_i$. 
We input the instance text $x_i^j$ to the LLM $f$ and perform post-processing on the answer returned by the LLM to obtain the final answer $\hat{y}_i^j$:
\begin{equation}
    \hat{y}_i^j = \text{PO}(f(x_i^j)).
\label{eq-1}
\end{equation}
where $\text{PO}$ represents $\textsc{P\scalebox{0.8}{OST}P\scalebox{0.8}{ROCESS}}$.
After obtaining the final answer $\hat{y}_i^j$, we can leverage evaluation metric to acquire the real performance $p_i$ of LLM $f$ on the task $T_i$:
\begin{equation}
    p_i = \frac{1}{|T_i|}{\sum_{j=1}^{|T_i|}\text{ME}(\hat{y}_i^j, y_i^j)}.
\label{eq-2}
\end{equation}
where $\text{ME}$ represents a $\text{M\scalebox{0.8}{ETRIC}}$, such as Exact Match Accuracy, ROUGE~\cite{lin2004rouge}, and so on.
By testing $f$ on all the tasks in the LLM Benchmark $T$, we can obtain an assessment of the LLM $f$'s capabilities on various tasks $\bm{p}=[p_1,\cdots,p_N]$. 
After evaluating all the LLMs in the model zoo, we can also obtain the ranking $\bm{r}=[r_1,\cdots,r_N]$ of the LLM $f$.
$\bm{p}$ and $\bm{r}$ respectively reflect the LLM's absolute and relative performance.
We need $k=\sum_{i=1}^{N}|T_i|$ forward inferences to evaluate the LLM $f$ on LLM Benchmark $T$. 
With each inference taking an average of $t$ seconds, the total evaluation time is $k\times t$. 
Given the numerous test instances in LLM Benchmark and the relatively slow inference speed of LLMs, both $k$ and $t$ tend to be high, significantly increasing the time and computational resources required for evaluation. Our work is aimed to decrease $k$.

\subsection{Efficient Evaluation Method for LLMs}\label{section:efficient_method}
In practical scenarios, different LLMs can be organized into an ordered list $[f_1,f_2,f_3,\cdots]$ based on their release dates. 
We assume that we can get the evaluation results $D\in\mathbb{R}^{B\times|T|}$ cosisting of $\text{ME}(\hat{y}_i^j, y_i^j)$ for the earliest released $B$ LLMs $[f_1, \cdots, f_B]$ on all instances of the LLM Benchmark. 
We refer to this collection of $B$ LLMs as the initial model set. 
Without loss of generality, we assume that larger values in $D$ correspond to higher quality answers from the LLM on the respective instances, indicating better performance, conversely, smaller values reflect lower answer quality.
We now focus on evaluating a new LLM on the Benchmark, mainly on its absolute performance and its ranking relative to the initial LLMs for each task.
The previous method tests the LLM on all instances in the Benchmark to obtain the ground truth values of model performance $p_i$ and rankings $r_i$, as shown in \autoref{eq-2}. 
While efficient evaluation method aims to obtain precise estimates of $p_i$ and $r_i$ by utilizing only a subset of instances on the Benchmark.

A well-designed efficient evaluation method consists of two core components: the \textbf{instance selection function} 
$g$ and the \textbf{performance prediction function} $h$.
The instance sampling function $g$ leverages the evaluation results of initial LLMs $D_i \in \mathbb{R}^{B\times|T_i|}$ on all instances for each task to select the important test instances, which forms a subset $T^s_i$ of $T_i$:
\begin{equation}
    T^s_i=g(D_i,T_i), \ |T^s_i|<|T_i|, \ T^s_i \subsetneqq T_i.
\label{eq-3}
\end{equation}
After getting the subset benchmark $T^s = \{T_1^s, \cdots, T_N^s\}$, the performance prediction function $h$ uses these subsets and the evaluation result of initial LLMs on the $i$-th task $D_i \in \mathbb{R}^{B\times|T_i|}$ to get estimate performance $\hat{p}_i$ of the real performance $p_i$ of the new model for each task.
\begin{equation}
    \hat{p}_i=h(D_i,T^s_i).
\label{eq-4}
\end{equation}
After getting $\hat{p}_i$, we can compare it with the performance of historical LLMs to determine its ranking $\hat{r}_i$ on each task.

Different efficient evaluation methods have different implementations of $g$ and $h$. To compare these methods, we use the Mean Absolute Error (MAE) between the LLM's predicted performances $\bm{\hat{p}}=[\hat{p}_1, \cdots, \hat{p}_N]$ and real performances $\bm{p}$ as shown in \autoref{eq-5}. The MAE between predicted rankings $\bm{\hat{r}}=[\hat{r}_1, \cdots, \hat{r}_N]$ and real rankings $\bm{r}$ across the tasks can be computed in a similar way.
\begin{equation}
\begin{aligned}
    \text{MAE}(\bm{p}, \bm{\hat{p}})=\frac{1}{N}{\sum_{i=1}^N |p_i-\hat{p}_i|}.
\end{aligned}
\label{eq-5}
\end{equation}
When comparing the MAE of different methods, we ensure that the numbers of selected instance subsets $|T^s|$ for different methods are consistent. 
A smaller MAE indicates better performance of the corresponding efficient evaluation method.
Note that we do not use Kendall's $\tau$ or weighted Kendall's $\tau$ coefficient because the ranking error of a new model will not cause a significant change in the above metrics, making it less effective for further comparisons.

When designing an efficient evaluation method, the following criteria should be met: 
$\textbf{1) High Efficiency}$. The method should give the predicted result accurately and quickly to satisfy the low-cost requirements mentioned in Section~\ref{section:1}. $\textbf{2) Low Overhead}$. The method should be easily deployed without additional overhead so that can be used quickly for new scenarios. $\textbf{3) Commonness}$. We need to quickly evaluate the overall capability of the model on a task with a limited number of samples, which means reducing the MAE of performances in \autoref{eq-5}. So, the method needs to reasonably handle instances with redundant information and keep only a small number of them.  $\textbf{4) Personalization}$. A good benchmark~\cite{2023opencompass, openllmleaderboard} should not only offer the absolute performance of models, but also provide their relative differences, such as their rankings $\bm{r}$. Hence, efficient evaluation methods should consider the differences of each model to give an accurate ranking between models. $\textbf{5) Complementation}$. Given that the capabilities evaluated across different tasks may sometimes be similar, a good method should use information from other tasks to improve their performance or speed.

\subsection{Insights from A Simple Baseline}\label{section:clustering}

In this subsection, we offer a simple implementation of an efficient evaluation method as a baseline. We also do a toy experiment and find that semantic information is not helpful.

In this baseline, we implement the instance selection function $g$ using the K-means clustering algorithm on instance embeddings. Specifically, we perform the clustering algorithm on each task and select instances closest to the cluster center as selected instances. 
For performance prediction function $h$, we use the 
following equation to get the predicted peformances $\hat{p}$:

\begin{equation}
    \hat{p}_i=\frac{\sum_{j=1}^K |C_{ij}|f(x_{ij})}{\sum_{j=1}^K|C_{ij}|}
\label{eq-6}
\end{equation}

where $|C_{ij}|$ is instance number of the $j$-the cluster in task $i$ and $x_{ij}$ is the instance colsest to the center of $c_{ij}$. This method assumes that the evaluation results of the instances of the same cluster are consistent with the center points and the weighted average of the center points can be used as an estimate of the real performance.

\begin{table}[t]
\centering
\vspace{-0.4cm}  
\caption{\textbf{Performance between Embedding Methods}.}
\label{table-embedding}
\small
\begin{tabular}{@{}m{3cm} m{2cm} m{2cm} m{1cm} m{1cm} m{1cm}@{}}
\toprule
\centering \textbf{Embedding Method} & \centering \textbf{MAE of Performance} & \centering \textbf{MAE of Ranking}\tabularnewline \midrule
\centering \textbf{Historical Results}       & \centering $0.035$ & \centering $2.9$\tabularnewline
\centering \textbf{Semantic}       & \centering $0.210$ & \centering$ 6.7$ \tabularnewline
\bottomrule
\end{tabular}
\end{table}

We implement two most common instance embedding methods for clustering, the first is semantic embedding given by Sentence-Bert~\cite{sentence_bert} and the other is the historical evaluation results of initial LLMs, which is the same as $D_i$ in Section~\ref{section:efficient_method}. 
We do a toy experiment on a subset of Opencompass Benchmark~\cite{2023opencompass} to compare two instance embedding methods and the result is in \autoref{table-embedding}, we can see that the efficient evaluation method with LLMs' historical evaluation results is better than that with semantic embedding. 
We guess the reason is that there is a large gap between the embedding space and the evaluation result space. Using historical evaluation results can better reflect the difficulty of samples than semantic information, so it is easier to predict the results of unselected instances. 
Experimental details and testing of hypotheses can be viewed in Appendix \ref{sec:appendixA}.

In addition to the poor performance of semantic embedding, there are other difficulties in using semantic information:
1) High Consumption of Semantic Embedding Models. At least one embedding model infers on all the instances causing high consumption.
2) Large Space to Store Semantic Embeddings. A lot of storage space is needed to store semantic embeddings compared to historical model performance.
3) Privacy Requirements for Benchmark. Some benchmark providers only provide a subset of the benchmark dataset. It is unfair and unsafe for model developers to provide their models to benchmark providers.
In summary, we do not use semantic information in our method but only the evaluation results of historical models as in the previous methods.

\section{Method}\label{section:our_methods}

This section presents a two-stage method based on Collaborative Filtering (CF) and highlights the advantages of our method over previous approaches. 
By treating LLMs as users and test instances as items, we construct user and item features from user-item interactions. 
These features help calculate similarities between new and previous LLMs, and between selected and unselected instances. 
These similarities are then used for instance selection and performance prediction. 
In addition, we use historical information of similar scenes to obtain extra information to improve the performance of our method, which can be seen as meta-information in cold start problems.

\subsection{Stage 1: Select Test Instances via CF}\label{section:select}

\begin{figure}[t]
\centering  
\subfigure{
\includegraphics[width=\linewidth]{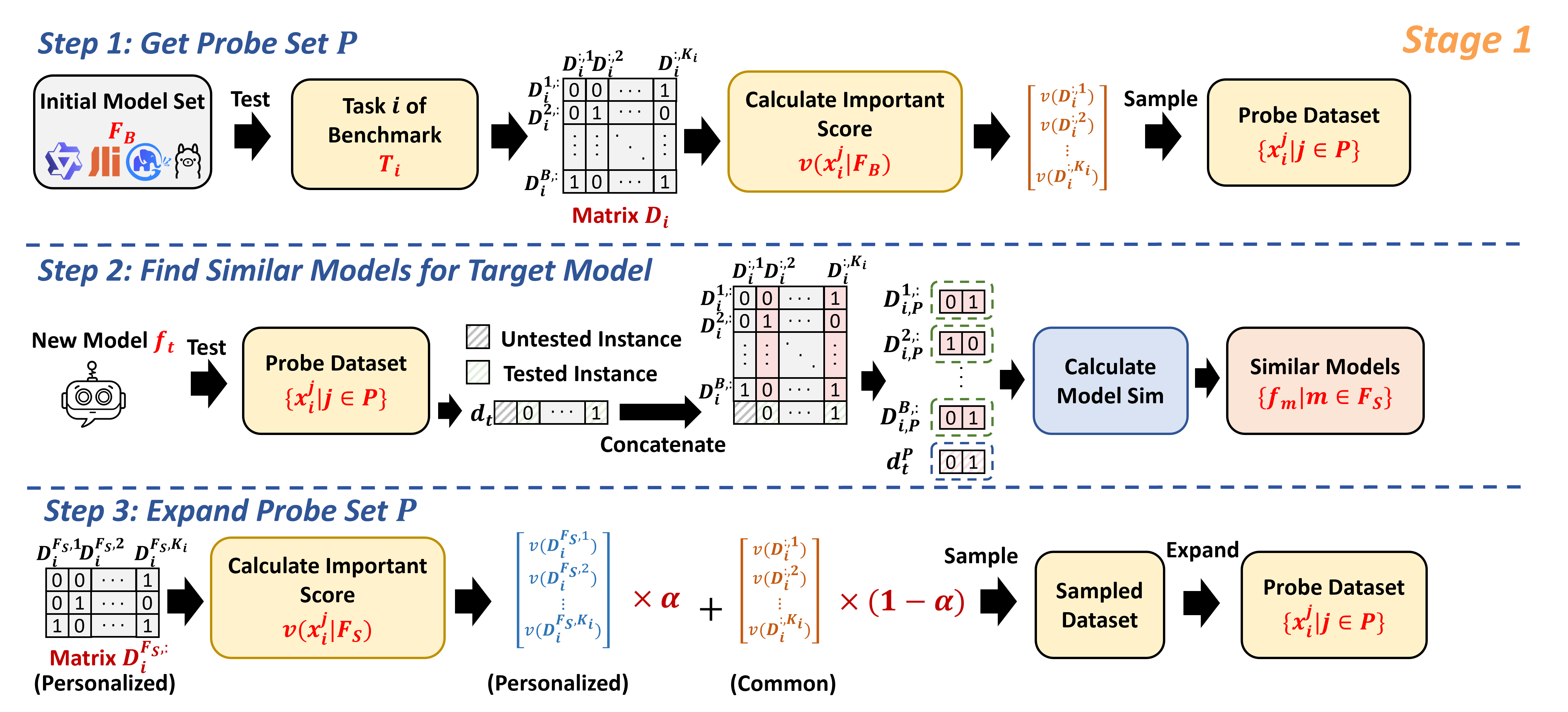}}
\vspace{-0.4cm}
\caption{\textbf{Steps in Instance Selection Process.} We select instances that can easily distinguish models through an iterative process.}
\label{fig:select}
\vspace{-0.4cm}
\end{figure}

As described in Section~\ref{section:efficient_method}, the instance selection function $g$ aims to select important instances for each task. 
We first define the importance score, and then design an iterative sampling process for each task as shown in \autoref{fig:select}.

\subsubsection{Definition of Important Score}\label{section:important_score}
We first define the importance score of a test instance for a model set. 
We think instances that can easily distinguish model performance are important.
This is consistent with research in educational measurement~\citep{CAT}, which shows that overly difficult or easy instances fail to differentiate between test takers of varying abilities, as all test takers may answer correctly or incorrectly. 
Intuitively, the most important test instances are those where half of the LLMs provide high-quality answers and the other half provide low-quality answers.
Such a design ensures that any LLM can be distinctly differentiated from half of the other LLMs in the model set on the current instance. We define the importance score $v(x|F)$ for instance $x$ given model set $F=\{f_1, \cdots, f_M\}$ as:

\begin{equation}
    v(x|F) = \\ \frac{1}{M - 1}{\sum_{m=1}^{M}(\text{ME}(\hat{y}^{(f_m)}, y) - \overline{\text{ME}}(\hat{y}, y))^2}.
\label{eq-7}
\end{equation}

where $\hat{y}^{(f_m)} = \text{PO}(f_m(x))$, $y$ and $\overline{\text{ME}}$  indicate the $m$-th LLM's answer, ground truth answer and average performance for instance $x$, respectively. 
The metric represents the variability in the LLM's responses for a single instance.
The larger this value is, the higher the importance score of the instance is.

\subsubsection{Process for
Instance Sampling}

To choose instances that really matter, the selected instances should meet two conditions:
(1) The instances should be able to assess the capabilities of all LLMs in the model set, meaning that the instances are required to have a high importance score $v(x|F_B \cup \{f_t\})$ for the model set $F_B \cup \{f_t\}$ composed of the $B$ initial LLMs $F_B$ and the target LLM $f_t$; 
(2) The instances should primarily differentiate the target LLM from its similar LLMs in $F_B$, which requires the instances to have a high importance score $v(x|F_S \cup \{f_t\})$ for the model set $F_S \cup \{f_t\}$ composed of the target LLM $f_t$ and it's similar LLMs $F_S$. 

To meet the above requirements, we design an iterative process. As \autoref{fig:select} shows, given target LLM $f_t$, the $i$-th task $T_i$, initial models $F_B=\{f_1, \cdots, f_B\}$, evaluation results of initial models $D_i \in \mathbb{R}^{B \times |T_i|}$, the process can select important instances with three steps.

\textbf{In the first step}, we get a Probe Set $P$ by calculating the important score for each instance in $T_i$ and selecting instances with higher important scores. For example, important score of the $j$-th instance in $T_i$ can be calculated as follows:
\begin{equation}
    v(x_i^j|F_B) = \frac{1}{B - 1}{\sum_{k=1}^{B}(D_i^{kj} - \overline{D_i^{\boldsymbol{\cdot} j}})^2}.
\label{eq-9}
\end{equation}
where $D_i^{kj}$ refers to the element in the $k$-th row and $j$-th column of $D_i$, which corresponds to the evaluation result of the $k$-th LLM on the $j$-th instance.
And $\overline{D_i^{\boldsymbol{\cdot} j}}$ represents the average of the $j$-th column of $D_i$.
This is to find instances that can easily distinguish performance between models.

\textbf{In the second step}, we let target model $f_t$ infer on the Probe Set $P$ to get the evaluation result vector $d_t^P \in \mathbb{R}^{|P|}$. 
Subsequently, we extract the evaluation results $D_{i,P} \in \mathbb{R}^{B\times |P|}$ of the initial LLMs $F_B$ on the instances in $P$ from $D_i$.
We treat $d_t^p$ and each row of $D_{i,P}$ as the features of the $f_t$ and $F_B$, respectively.
By calculating the cosine similarity between each row of $D_{i,P}$ and $d_t^p$, and selecting the top $n$ LLMs most similar to the target LLM $f_t$, we can form a set of LLMs $F_S$ with capabilities similar to the current LLM.
We use $S = \{k|f_k \in F_S\}$ to represent the index set of similar LLMs.
We calculate the importance score of the instance $x_i^j$ on the similar model set $F_S$:
\begin{equation}
    v(x_i^j|F_S) = \frac{1}{\left| S \right| - 1}{\sum_{k\in{S}}(D_i^{kj} - \overline{D_i^{\cdot j}})^2}.
\label{eq-10}
\end{equation}
This is to select instances that should primarily differentiate the target LLM from its similar LLMs in $F_B$.

\textbf{In the third step}, we perform a weighted sum of $v(x_i^j|F_B)$ and $v(x_i^j|F_S)$ like \autoref{eq-11} to obtain the instance's final importance score $v(x_i^j)$ thus meeting two conditions talked above. Then we select new $q$ instances with higher $v(x_i^j)$ to expand the Probe Set $P$.
\begin{equation}
    v(x_i^j) = \alpha \times v(x_i^j|F_B) + (1-\alpha) \times v(x_i^j|F_S).
\label{eq-11}
\end{equation}
Finally, we can repeat step 2 and 3 until $|P|$ meets the expected subset size $|T_i^s|$.
In \autoref{eq-11}, the first term remains constant across different target LLMs, while the second term varies because the sets of similar LLMs $F_S$ differ among different target LLMs. Hence, it achieves personalized instance selection.

\subsubsection{Relationship with Recommend Systems}

The process is similar to user-based collaborative filtering, where LLMs and instances act as users and items in a recommendation system, respectively. 
The evaluation matrix $D_B$ of LLMs on instances is similar to the user-item interactions matrix.
The difference is that the values in our matrix reflect the quality of LLM responses, not user preferences. 
Our goal is to recommend the most important instances (items) to a new LLM (user). 
Our method starts by testing the target LLM on a Probe Set, like recommending popular items to a new user in cold start problems.
Our instances sampling process resembles the user-based collaborative filtering while also considering the popularity of products to recommend items to new users.

\subsection{Stage 2: Predict LLM's Performance}\label{section:predict}

After getting the subset benchmark $T^s = \{T_1^s, \cdots, T_N^s\}$, we need to design the performance prediction function $h$ to predict the performance $p_i$ and ranking $r_i$ of the new LLM. 

\subsubsection{Purpose of Performance Prediction}

For performance prediction, a direct method is to use the new LLM's performance $p_i^s$ on the sampled subset $T_i^s$. 
However, since our instance selecting method excludes those that are either too difficult or too easy, $p_i^s$ may not be an exact estimate of $p_i$.
To obtain an accurate estimate of $p_i$, we reexamine each component of its calculation formula~(\autoref{eq-12}).
We use $T_i^s$ and $T_i^{ns}$ to represent the instances set we have selected and not selected, respectively.
Similarly, $a_i^s$ and $a_i^{ns}$ denote the sum of the real performances of the target LLM $f_t$ for instances on sets $T_i^s$ and $T_i^{ns}$.
\begin{equation}
    p_i = \frac{a_i^s + a_i^{ns}}{|T_i^s| + |T_i^{ns}|}.
\label{eq-12}
\end{equation}
Since we can obtain the true value of $|T_i^s|$, $|T_i^{ns}|$ and $a_i^s$, we only need to predict the value of $a_i^{ns}$ to get $p_i$. 

\subsubsection{Process For Performance Prediction}

However, $|T_i^s|$ is usually small making predicting the value of $a_i^{ns}$ difficult. Fortunately, this is similar to the cold start problem in recommendation systems, and we decide to draw inspiration from the idea of using meta-information to assist recommendations.  
However, in light of the issues mentioned in Section~\ref{section:clustering}, we choose not to rely on semantic information.  
Instead, we leverage the performance of sampled instances in similar tasks combined with Optimal Transport~(OT) to derive meta-information, in other words, synthetic data, thereby enhancing evaluation performance.

\begin{figure}[t]
\centering  
\subfigure{
\includegraphics[width=0.9\linewidth]{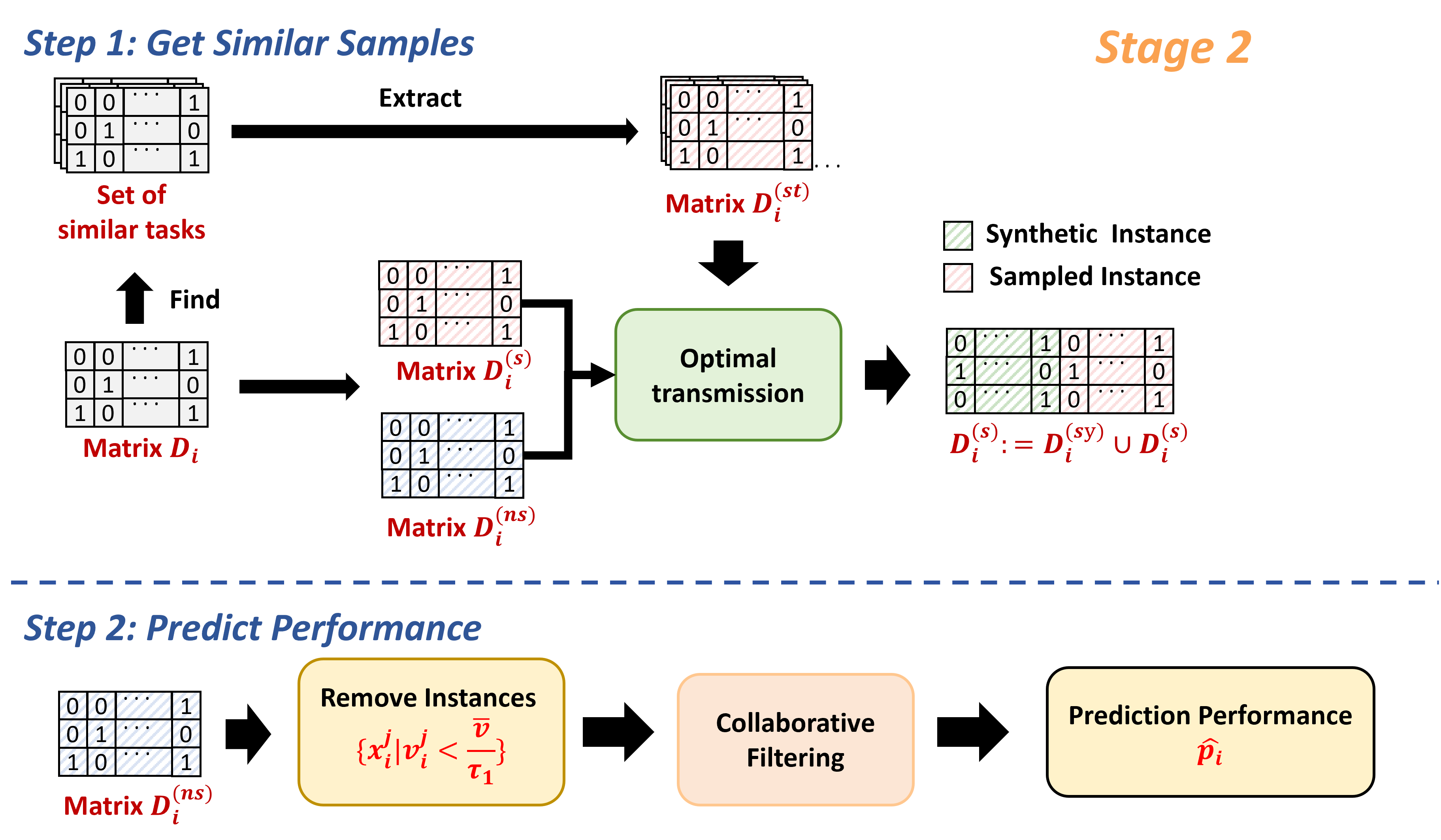}}
\vspace{-0.4cm}
\caption{\textbf{Steps in Performance Prediction Process.} We predict performance based on optimal transport and collaborative filtering.}
\label{fig:predict}
\vspace{-0.4cm}
\end{figure}

\textbf{For the first step} in \autoref{fig:predict}, for the $i$-th task, supposing the evaluation results of initial LLMs $D_i \in \mathbb{R}^{B \times |T_i|}$ and the LLM $f_t$ on sampled dataset $d_t = [\text{ME}(\hat{y}_i^1,y_i^1), \cdots, \text{ME}(\hat{y}_i^{|T_i^s|},y_i^{|T_i^s|})] \in \mathbb{R}^{1\times|T_i^s|}$.
We can get the mean performance vector $\mathbf{v}_i = \frac{\sum_{j=1}^{|T_i|}D_i^{\cdot j}}{|T_i|}$ of initial LLMs for each task and calculate cosine similarity matrix $C_s\in \mathbb{R}^{N\times N}$. With the help of $C_s$ and threshold value $\tau_0$, similar tasks can be found and we can combine the selected data from similar tasks to get $T_i^{st}$ and $D_i^{(st)} \in \mathbb{R}^{B \times |T_i^{st}|}$. We can also extract a submatrix $D_i^{(ns)} \in \mathbb{R}^{B \times |T_i^{ns}|}$ which represents unselected samples from $D_i$.
Then we can do the Optimal Transport~(OT) problem as follows:
\begin{equation}
\begin{aligned}
& \operatorname*{\text{argmin}}_{P \geq 0} && \langle C_c, P \rangle \\
& \text{subject to} && \sum_{j=1}^{|T_i^{st}|} P_{jk} = \frac{1}{|T_i^{ns}|}, \\
& && \sum_{k=1}^{|T_i^{ns}|} P_{jk} = \frac{1}{|T_i^{st}|}
\end{aligned}
\end{equation}
\label{eq-13}

where $C_c \in \mathbb{R}^{|T_i^{st}| \times |T_i^{ns}|}$ is cost matrix between columns of $D_i^{(st)}$ and columns of $D_i^{(ns)}$ caculated by euler distance. With $P$, we can get synthetic dataset $T_i^{sy}$ and $D_i^{(sy)} \in \mathbb{R}^{B \times |T_i^{ns}|}$, where $D_i^{(sy)}=D_i^{(st)}P$. Because the real value of $f_t$ on $T_i^s$, the value of $f_t$ on the synthetic dataset can be calculated in the same way.

Now come to \textbf{the second step} in \autoref{fig:predict}.
For the sake of simplicity, we treat the synthetic data just like sampled instance, that is to say, $T_i := T_i^s \cup T_i^{ns} \cup T_i^{sy}$ and $T_i^s := T_i^s\cup T_i^{sy}$.
We denote the matrices composed of feature vectors of selected and unselected instances as $D_i^{(s)} \in \mathbb{R}^{B\times|T_i^s|}$ and $D_i^{(ns)} \in \mathbb{R}^{B\times|T_i^{ns}|}$.
We then calculate the average value of importance scores $\overline{v}$ use \autoref{eq-7} for $D_i^{(s)}$ and importance score for each instance in $D_i^{(ns)}$.
We set a threshold $\tau_1$. 
For instances whose importance score is smaller than $\frac{\overline{v}}{\tau_1}$, we use the average performance of init LLMs as predicted performance $\hat{c}_{i0}^{ns}$ and remove them from $D_i^{(ns)}$
:
\begin{equation}
\begin{aligned}
    \hat{c}_{i0}^{ns} &= \frac{1}{B}{\sum_{k=1}^B\left(\sum_{j\in S}D_i^{(ns)kj}\right)}.\\
    T_i^{ns} &:= T_i^{ns} - S\\
    D_i^{(ns)} &:= D_i^{(ns)} - S \in \mathbb{R}^{B \times (|T_i^{(ns)}| - |S|)}
\end{aligned}
\label{eq-14}
\end{equation}
where $S=\{k| v_i^{(ns)k} < \frac{\overline{v}}{\tau_1}\}$. 
As discussed in Section~\ref{section:important_score}, We do this because we believe that too difficult or easy instances fail to differentiate models. 

After that, we can calculate the cosine similarity matrix $C_i\in \mathbb{R}^{|T_i^{ns}|\times|T_i^{s}|}$ between the selected instances' feature $D_i^{(s)}$ and unselected instances $D_i^{(ns)}$ for each task.
And for each unselected instance, we find the top 3 most similar instances and compute their average similarity $\overline{c}_i^j$. 
We set an adaptive threshold $\tau_3$. 
Based on the average similarity $\overline{c}_i^j$ and $\tau_3$, the instances can be divided into two parts.

For each instance in $\{x_i^j|\overline{c}_i^j \geq \tau_3\}$, we use the average evaluation results of the target LLM $f_t$ on the three most similar sampled instances as the predicted performance.
Hence, we can get the estimated sum of $f_t$'s performances $\hat{c}_{i1}^{ns}$ on $\{x_i^j|\overline{c}_i^j \geq \tau_3\}$.
This method is similar to the idea of item-based collaborative filtering.

\begin{figure*}[t]
    \centering  
    \includegraphics[width=\linewidth]{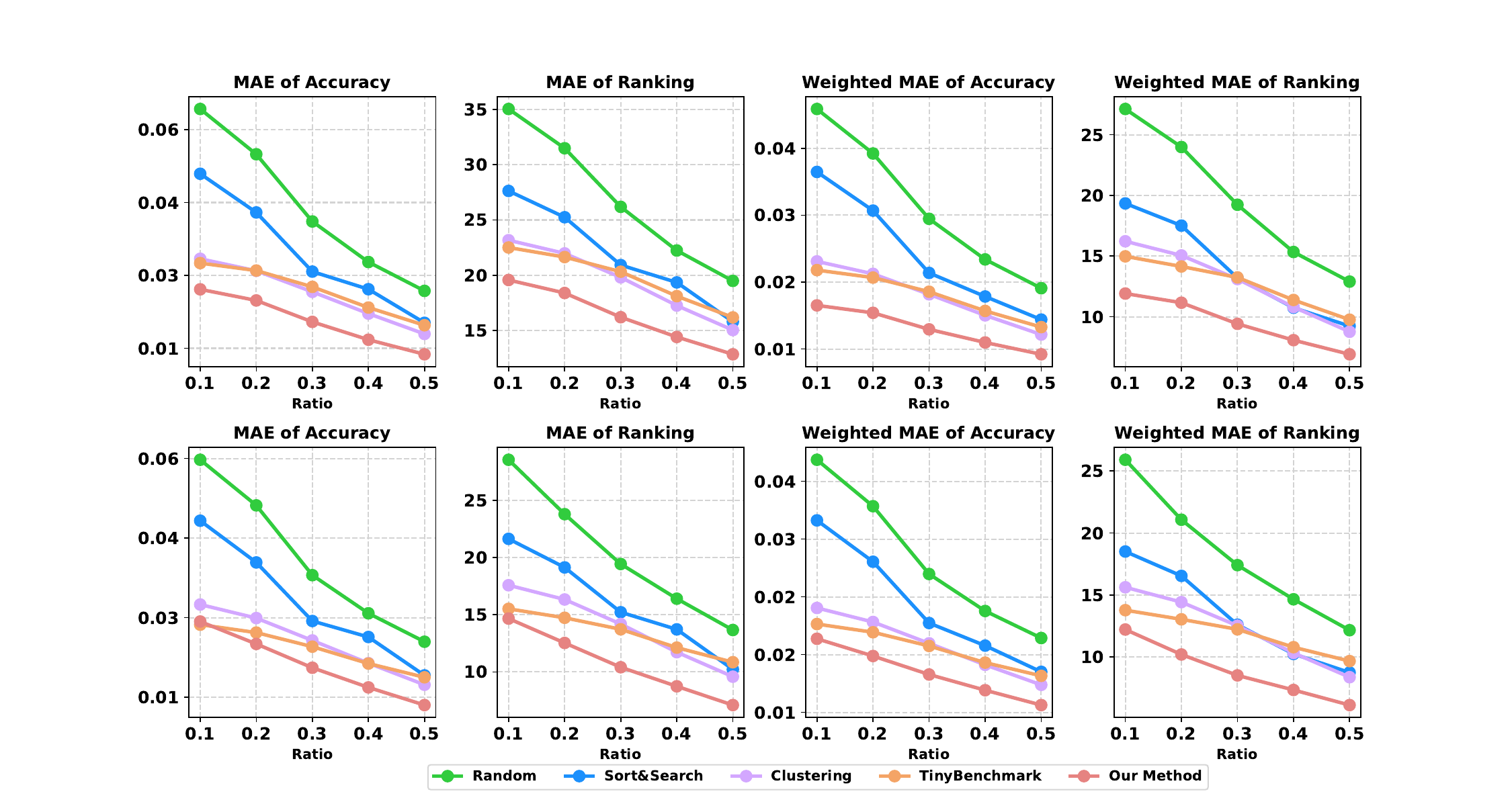} 
    \vspace{-0.6cm}
    \caption{The Mean Absolute Error~(MAE) and weighted MAE between the estimated LLM's performance by different efficient evaluation methods and the real performance of LLMs. \textbf{The first and second lines represent the results on MMLU and LB, respectively.}}
    \label{fig-exp-1}
    \vspace{-0.4cm}
\end{figure*}
\label{sec:performace_mae}

For each instance in $\{x_i^j|\overline{c}_i^j < \tau_3\}$, we use the average results of $f_t$'s similar LLMs as the predicted performance. And we can get the estimated sum of $f_t$'s performances $\hat{c}_{i2}^{ns}$.
This is similar to user-based collaborative filtering.

Thus, we can replace $a_i^{ns}$ with $\hat{c}_i^{ns}=\hat{c}_{i0}^{ns}+\hat{c}_{i1}^{ns}+\hat{c}_{i2}^{ns}$ in \autoref{eq-12} to obtain an estimate of model performance $\hat{p}_i$. Through experiments, we find that $\hat{p}_i$ yields more precise results compared to $p_i^s$. 

For LLM's ranking prediction, we compare the estimated performance $\hat{p}_i$ of the target LLM $f_t$ with the performances of initial LLMs to obtain the ranking prediction $\hat{r}_i$.

\subsection{Comparison with Previous Methods}
\begin{table}[t]
\centering
\vspace{-0.4cm}  
\caption{\textbf{Comparison of Different Evaluation Methods}.}
\label{tab:comparison}
\small
\begin{tabular}{@{}m{2.5cm} m{0.8cm} m{0.8cm} m{0.8cm} m{1.4cm} m{2cm}@{}}
\toprule
\centering \textbf{Method} &\centering  \textbf{Ours} & \centering \textbf{Cluster-ing} & \centering \textbf{Sort\&\newline Search} &\centering  \textbf{Tiny\newline Benchmark}\tabularnewline \midrule
\centering \textbf{High Efficiency}       & \centering \textcolor{mygreen}{\checkmark} &\centering  \textcolor{mygreen}{\checkmark} & \centering \textcolor{mygreen}{\checkmark} & \centering \textcolor{mygreen}{\checkmark}\tabularnewline
\centering \textbf{Low Overhead}       &\centering  \textcolor{mygreen}{\checkmark} &\centering  \textcolor{mygreen}{\checkmark} & \centering \textcolor{mygreen}{\checkmark} &\centering  \textcolor{myred}{\xmark}\tabularnewline
\centering \textbf{Commonness}           &\centering  \textcolor{mygreen}{\checkmark} &\centering  \textcolor{mygreen}{\checkmark} &\centering  \textcolor{mygreen}{\checkmark} &\centering  \textcolor{mygreen}{\checkmark}\tabularnewline
\centering \textbf{Personalization}     &\centering  \textcolor{mygreen}{\checkmark} &\centering  \textcolor{myred}{\xmark} & \centering \textcolor{mygreen}{\checkmark} &\centering  \textcolor{myred}{\xmark}\tabularnewline
\centering \textbf{Complementation}     &\centering  \textcolor{mygreen}{\checkmark} &\centering  \textcolor{myred}{\xmark} & \centering \textcolor{myred}{\xmark} & \centering \textcolor{mygreen}{\checkmark}\tabularnewline
\bottomrule
\end{tabular}
\end{table}

As shown in \autoref{tab:comparison}, our method meets all the criteria described in Section~\ref{section:efficient_method} which was not achievable with previous approaches. Specifically, our method only requires instances that are discriminative for models, which makes it \textbf{efficient} and \textbf{personalized}. For instances that are too simple or too difficult, our method uses the average results of the initial models for estimation. This ensures that our approach meets the \textbf{commonness} criterion. In addition, our method uses OT to leverage information from other tasks for prediction, thus meeting the \textbf{complementation} criterion.

\section{Experiments}\label{section:experiments}

\textbf{Setups.} 
Our experiment mainly focuses on efficiently evaluating the performance of new LLMs on Benchmarks, given the evaluation results of some initial LLMs on the Benchmark datasets. 
This aligns with real-world scenarios.
Based on their release dates, we select some of the earliest released LLMs for each Benchmark as initial LLMs.
As described in Section~\ref{section:1}, the overall performance of an LLM cannot reflect its task-level performance. Therefore, we focus on whether the efficient evaluation method can accurately predict the LLM's performance on each task.
To compare the adaptability of the efficient evaluation method to different sample sizes, we set 5 ratios~([0.1, 0.2, 0.3, 0.4, 0.5]) to select corresponding subsets from each task's dataset.

\textbf{Benchmark.} 
We choose two commonly used LLM Benchmarks for our experiments. 
(1) HuggingFace’s Open LLM Leaderboard is a public ranking platform designed to compare and showcase the performance of open-source large language models (LLMs). It provides a standardized evaluation framework across multiple benchmarks, including tasks like natural language understanding, generation, and reasoning. We follow TinyBenchmark~\citep{tiny_benchmark} to collect the evaluation results of 395 LLMs and get approximately 50K instances.
We divide the 395 LLMs into initial LLMs and test LLMs based on their release dates in a 3:1 ratio.
(2) MMLU: MMLU is a multiple-choice QA benchmark consisting of 57 tasks. It comprises approximately 14K instances and we consider the same set of 395 LLMs and train-test splits. 
The reason to consider it is its immense popularity when comparing LLMs and inclusion into several other benchmarks.

\textbf{Baselines.} We adopt four baselines as our comparison methods, which are Random Sampling, Baseline with Clustering, TinyBenchmark, and Sort\&Search.
Since Baseline with Clustering has been introduced in Section~\ref{section:clustering}, we do not repeat it here.
For Random Sampling, we randomly sample instances from each task's dataset to form its subset. 
Then we use the LLM's performance on these subsets to act as the estimation for its performance on each task's dataset. 
We rank different LLMs based on their performances on the sampled subsets. 
For TinyBenchmark~\citep{tiny_benchmark} and Sort\&Search~\citep{lifelong_benchmark}, we compare the target LLM's estimated performance with the real performance of other LLMs on each task's dataset to obtain the ranking of the target LLM.
For methods with randomness, we repeat the above experiment 5 times and report the mean as result.

\textbf{Evaluations.} 
For each task, we calculate the Mean Absolute Error~(MAE) between the predicted performance and the actual performance for each model.
Then, we take the mean of MAE for all tasks as the final performance of the Efficient Benchmark Method.
For example, for the MMLU Benchmark, we should calculate the average MAE for 395 test models on 57 datasets, totaling 22,515 test cases.
Additionally, considering that we are more concerned with evaluating strong models in practice, we calculate the weighted MAE by using the scaling factor $\frac{1}{r_i}$ to weight the MAE of different LLMs based on their true rankings $r_i$. 

\textbf{Results Analysis.} 
\autoref{fig-exp-1} shows the MAE and weighted MAE between the estimated LLM's performance metrics~(performance and ranking) by different Efficient Benchmark methods and the actual performance metrics of LLMs.
The smaller the values of MAE and Weighted MAE, the more accurate the performance metrics of LLMs estimated by Efficient Benchmark are.
From the \autoref{fig-exp-1}, we can see that on benchmarks called Open LLM
Leaderboard and MMLU our method obtains the most accurate estimates of LLM's performance. 
Moreover, surprisingly, Baseline with Clustering performs very well, and only our method consistently outperforms it across different ratios.

\textbf{Running Time Analysis}
The method's runtime is also an important metric to compare efficient evaluation methods. For simplicity, we assume that the inference time of a LLM is the same for different instances.
We show the time taken by different methods to evaluate 1 model on MMLU at a sampling ratio of 0.1. 
\autoref{table-1} shows the results, which include the time for methods to deploy (Deployment Time), the time to select instances (Select Time), the time to predict performance (Predict Time) and the sum of Select time and Predict Time (Total Time). 
Compared to TinyBenchmark and Clustering Baseline, our method demonstrates a smaller time cost while providing accurate LLM performance estimates. Although our method is slower than the Random and Sort\&Search methods, the slight increase in time is acceptable considering their poor performance.

\textbf{Ablation Study}
We conduct an ablation study on the components and parameters of the method in Section~\ref{section:our_methods}. The specific roles and details of the parameters can be found in Appendix \ref{sec:appendixB}.
The experiments demonstrate that optimal transport is an indispensable module in our method, and our method is robust to different parameter settings. Furthermore, it shows that the performance reported in Section \ref{section:experiments} is not the best, indicating that our method has further potential.

\begin{table}[t]
\centering
\vspace{-0.2cm}
\caption{\textbf{Comparison of Time Between Methods}.}
\label{table-1}
\small
\begin{tabular}{@{}m{2cm} m{1cm} m{1cm} m{1cm} m{1cm} m{1cm}@{}}
\toprule
\centering \textbf{Method} & \centering \textbf{DT(s)} & \centering \textbf{ST(s)} & \centering \textbf{PT(s)}& \centering \textbf{TT(s)}\tabularnewline \midrule
\centering \textbf{Random}       & \centering 0 & \centering $0.00149$ & \centering $0.000782$ & \centering $0.00227$\tabularnewline
\centering \textbf{Sort\&Search}       & \centering 0 & \centering$ 6.05$ &\centering  $0.0167$ & \centering $6.07$ \tabularnewline
\centering \textbf{Clustering}           & \centering 0& \centering $26.8 $ &  \centering $0.232$ & \centering $27.0$\tabularnewline
\centering \textbf{TinyBenchmark}     & \centering $309$ & \centering $26.8$ & \centering $3.36$ & \centering $30.2$ \tabularnewline
\centering \textbf{Ours}     & \centering 0 & \centering $0.713$ & \centering $24.5 $& \centering $25.2$\tabularnewline
\bottomrule
\end{tabular}
\end{table}

\section{Conclusion}
In this work, we focus on designing an efficient evaluation method to evaluate the target LLM's task-level capacities at a low cost. 
We re-examine the issue from the perspectives of collaborative filtering in recommendation systems and propose a two-stage method, which includes instance selection stage and performance estimation stage. 
The experimental results across multiple LLMs and datasets demonstrate the effectiveness of our method.

\small
\bibliography{reference}
\bibliographystyle{plainnat}

\newpage
\appendix
\onecolumn

In the Appendix, we introduce more details about the Experiments.

\section{Experiments for Clustering Baseline}\label{sec:appendixA}

\subsection{Experiment Settings}
 We sample 187 tasks~\cite{HellaSwag,arc,piqa,obqa,Race,fewclue,clue,ceval} from OpenCompass~\cite{2023opencompass}, a large language model evaluation benchmark. 
 We collect the evaluation results of 32 widely used LLMs~(\eg, LLAMA~\citep{llama}, Qwen~\citep{qwen}, ChatGLM~\citep{chatglm}, Gemma~\citep{gemma}). We select the first 21 LLMs based on their release dates as the initial LLMs and use the remaining 11 LLMs as the new LLMs to be tested.

We do the toy experiment with ratio 0.1, which is described in Section \ref{section:experiments}.

\subsection{Hypotheses Testing}\label{sec:appendix A.2}

\begin{figure*}[ht]
    \vspace{-0.3cm}
    \centering  
    \includegraphics[width=0.5\linewidth]{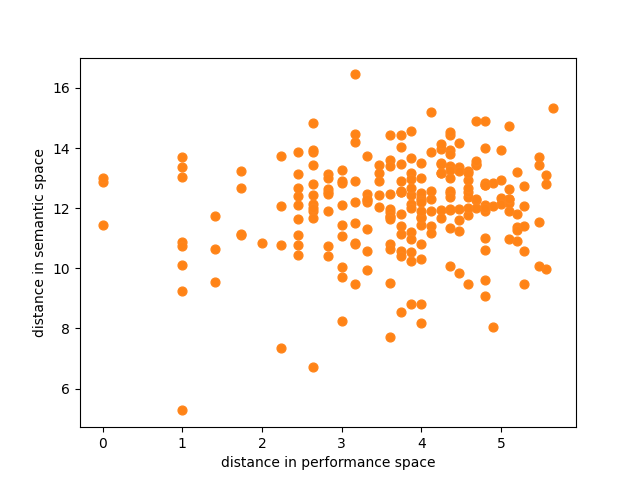} 
    \caption{\textbf{Distances in Two Spaces for Nearest Neighbor in Semantic Space.} Each point represents a nearest-neighbor pair in the semantic space. The vertical coordinate indicates the distance between the nearest neighbors in the semantic embedding space, while the horizontal coordinate represents the distance between the corresponding samples in the evaluation results space.}
    \label{fig:diff}
\end{figure*}

To verify the hypothesis that there is a gap between the semantic embedding space and the evaluation results space. We plot \autoref{fig:diff} on one task of MMLU benchmark.

We can see that there is no obvious relationship between the two distances meaning that the two spaces are not aligned.

\section{Something for Binary Evaluation Matrix}

If the ME in Section~\ref{section:original_evaluation} yields binary results of 0 or 1, in other words, the evaluation results matrix $D_i$ in Section~\ref{section:efficient_method} is a binary matrix, then some equations can be calculated using alternative methods or may require further steps.

\subsection{Another Way to Calculate Important Score}
With the ME yields binary results of 0 or 1, we can alternatively calculate $v(x|F)$ in Section~\ref{section:select} using the quantity difference between 0 and 1 values shown in \autoref{eq-another_important_score}.

\begin{equation}
v(x|F) = \\ \frac{1}{|\sum_{m=1}^M [\mathbb{I}(\hat{y}^{(f_m)}=y) - \mathbb{I}(\hat{y}^{(f_m)}\neq y)]|}.
\label{eq-another_important_score}
\end{equation}
where $\mathbb{I}(\cdot)$ denotes the indicator function, which takes the value 1 if the condition inside the parentheses is true, and 0 otherwise.

\subsection{Further steps for Optimal Transport}

If $D_i$ in Section~\ref{section:predict} is a binary matrix, further steps should be done after getting matrix $D_i^{(sy)}$ in Section~\ref{section:predict}. Specifically, the elements in $D_i^{(sy)}$ should be divided into 0-1 by 0.5 as the threshold.

\section{Instance Sampling Details}
Assume $|T_i|$ is the number of test instance in the dataset for the $i$-th task, and let $\alpha$ be the sampling ratio, then $\alpha \times |T_i|$ represents the number of sampled instances.
To ensure an accurate estimation of LLM performance, we set a minimum sample size of 20 for each dataset.
When $|T_i|$ is less than 20, we will use all samples from the current dataset. 
When $|T_i|$ is greater than or equal to 20 but $\alpha \times |T_i|$ is less than 20, we will set the number of sampled instances for the current dataset to 20.

\section{Ablation Study}\label{sec:appendixB}

\subsection{Optimal Transport Module}

\begin{figure*}[ht]
    \vspace{-0.3cm}
    \centering  
    \includegraphics[width=\linewidth]{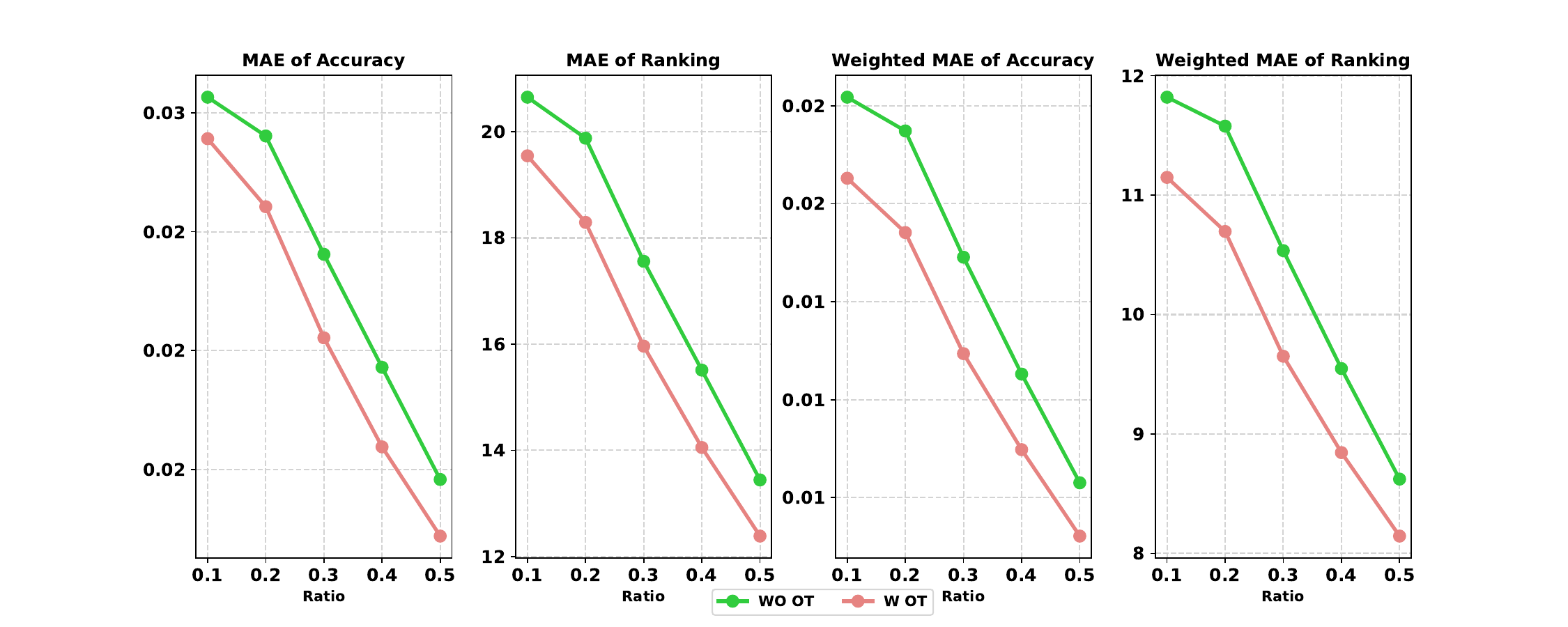} 
    \vspace{-1.0cm}
    \caption{\textbf{The MAE and Weighted MAE of Method with OT and Method without OT.}}
    \label{fig:ot}
\end{figure*}

We plot the MAE metric on different ratios on MMLU. \autoref{fig:ot} shows the result that our method with optimal transport is consistently better than that without optimal transport. All in all, introducing information from other tasks into efficient evaluation methods can improve method performance.

\subsection{Hyperparameter}

Next, we will analyze the hyperparameters involved in the method in Section~\ref{section:our_methods}. They are $|S|$, $\alpha$, number of iterations in Section~\ref{section:select} and $\tau_0$, $\tau_1$, $\tau_2$, $q$ in Section~\ref{section:predict}. Here $\tau_2$ and $q$ are hyperparameters related to $\tau_3$ in Section~\ref{section:predict}. Specifically, $\tau_3 = \text{max}(\tau_2, [\overline{\mathbf{c}}_{i}]_{q})$. 
Here $[\overline{\mathbf{c}}_{i}]_{q}$ refers to the $q$ quantile of $\overline{\mathbf{c}}_{i}$ and $\overline{\mathbf{c}}_{i}$ a vector formed by the average similarities of different tasks.

The specific roles of these hyperparameters are shown in \autoref{table-hypermarter}.

\begin{table}[ht]
\centering
\vspace{-0.2cm}
\caption{\textbf{Roles of Different Hyperparameters.}}
\label{table-hypermarter}
\adjustbox{max width=\textwidth}{ 
\small
\begin{tabular}{@{}p{4cm} p{5cm} p{5cm} p{2cm} p{2cm} p{2cm}@{}}
\toprule
\centering\textbf{Hyperparameter} & \centering\textbf{Roles}\tabularnewline \midrule
\centering \textbf{$|S|$}       & the number of similar models for a new target model \\
\centering \textbf{$\alpha$}       & a hyperparameter that determines the importance score of the sample\\
\centering number of iterations       & the number of iterations in instance selection process\\
\centering \textbf{$\tau_0$}           & a hyperparameter used to select similar tasks \\
\centering \textbf{$\tau_1$}     &  a hyperparameter used to determine unimportant instances\\
\centering \textbf{$\tau_2$}     & a hyperparameter used to determine the prediction mode \\
\centering \textbf{$q$}     & a hyperparameter used to determine the prediction mode \\
\bottomrule
\end{tabular}
} 
\end{table}

\begin{figure*}[ht]
    \vspace{-0.3cm}
    \centering  
    \includegraphics[width=\linewidth]{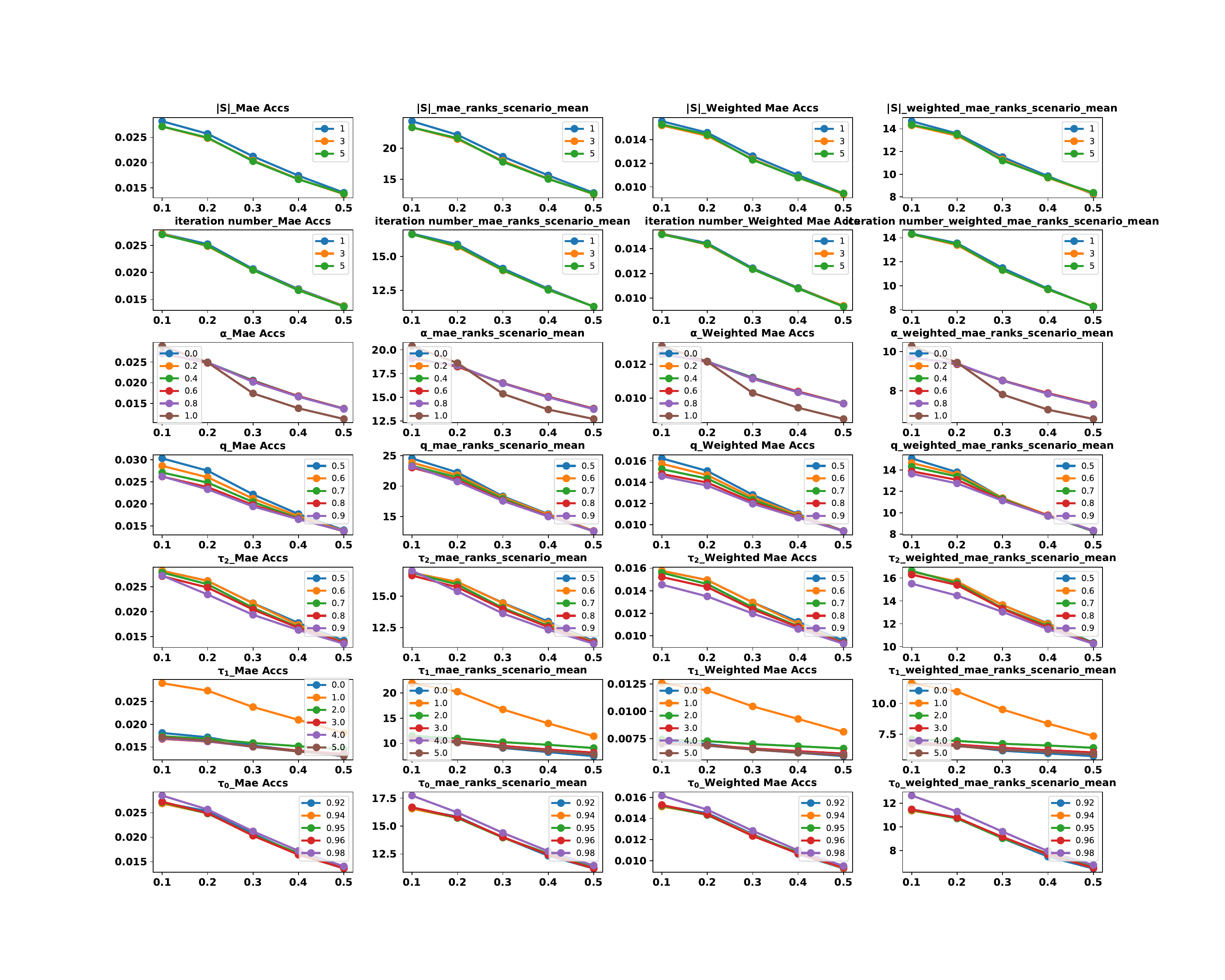} 
    \vspace{-1.0cm}
    \caption{\textbf{The MAE and Weighted MAE with Different Hyperparameters.}}
    \label{fig:abu_stu}
\end{figure*}

\autoref{fig:abu_stu} shows the MAE and weighted MAE with different hyperparameters on MMLU benchmark. Each row of the figure represents the performance change if only one hyperparameter is changed and the rest is unchanged. From top to bottom, they represent the $|S|$, number of iterations, $\alpha$, $q$, $\tau_2$, $\tau_1$ and $\tau_0$, respectively. From the figure, we can find that our method is robust to different hyperparameters.

\end{document}